\documentclass[10pt,twocolumn]{article}
\usepackage[letterpaper,margin=0.72in,columnsep=0.25in]{geometry}
\usepackage[T1]{fontenc}
\usepackage{lmodern,microtype,booktabs,graphicx,amsmath,xcolor,url,float,balance,seqsplit}
\usepackage[font=small,labelfont=bf]{caption}
\usepackage[round]{natbib}
\usepackage[colorlinks=true,allcolors=blue!45!black]{hyperref}
\hypersetup{pdftitle={Native Probability Outputs for Clinical Triage: A Benchmark Evaluation of Calibration and Decision Accuracy},pdfauthor={Liang You, Zhiyin Zhou, Dawei Tang, Chuchu Wu, Jingzhuo Hu},pdfsubject={Evaluation of a probabilistic decision model on patient-facing medical triage}}
\newcommand{\info}{\text{Info}}
\title{\vspace{-1em}Native Probability Outputs for Clinical Triage:
An Evaluation of Calibration and Decision Accuracy}

\author{
Liang You\\
\small University of Pittsburgh
\and
Zhiyin Zhou\\
\small Pratt Institute\\
\and
Dawei Tang\\
\small Icahn School of Medicine at Mount Sinai
\and
Chuchu Wu\\
\small Carnegie Mellon University
\and
Jingzhuo Hu\\
\small University of Pennsylvania
}
\date{September 21, 2026}
\begin{document}
\maketitle
\begin{abstract}\small
Patient-facing triage requires an appropriate care recommendation and recognition of when information is insufficient. We evaluate whether Jev, a decision model that returns probabilities over predefined actions, supports both judgments. Using CARE-Bench, we compare Jev with four general-purpose language models and three input-independent baselines across 925 decision points from 439 public cases. On the public test subset (135 decision points from 62 cases), Jev ranks fourth of five models by macro-F1 (0.461) but has the lowest observed multiclass Brier score (0.610), an 18.7\% improvement over uniform prediction (case-bootstrap 95\% interval, 4.9--31.1\%). Its top-label expected calibration error is 0.226, and it identifies only 3 of 37 decision points requiring clarification. In a follow-up that asks about information sufficiency separately, matched binary Choice and Noul interfaces each identify 36 of 37 such points, with 54 and 53 false positives among the remaining 98 points. Noul has lower binary Brier than Choice (0.215 versus 0.276). A secondary evaluation of 453 HealthBench conversations again gives Jev the lowest observed Brier score (0.299), but measures a distinct emergency-referral judgment. Jev combines lower probability error with missed clarification needs in the four-action task. Asking about information sufficiency separately increases recall but also produces frequent unnecessary clarification assignments.
\end{abstract}

\section{Introduction}
Patient-facing triage requires two judgments: what care is appropriate, and whether the available information supports that recommendation. A model may recognize an urgent presentation yet recommend care prematurely when essential details are missing. Evaluating the selected action alone also leaves a separate question unanswered: whether the model's probabilities distinguish decisions it is likely to get right from those it is likely to get wrong.

Jev returns probabilities over developer-defined actions, allowing action selection and probability quality to be evaluated together. This interface removes the need to infer an action from an open-ended response. The central question is whether these probabilities support appropriate care selection and recognition of insufficient information in patient-facing triage.

We examine three questions using CARE-Bench: How well does Jev select triage actions relative to general-purpose language models? How informative and calibrated are its probabilities? How does clarification recognition change when information sufficiency is assessed separately from care selection? A secondary analysis of HealthBench tests a related emergency-referral judgment after the full conversation (Section~4.4). We assess action selection, probability quality, calibration, and recognition of missing information. It also tests whether the apparent benefit of a binary sufficiency question is specific to Jev's Noul interface.

\section{Background}
TypeSafe describes Jev as trained with Reinforcement Learning for Calibrated Decisions (RLCD), an objective focused on decisions and their probabilities \citep{typesafeprimer}. Its Choice interface returns probabilities over named options, whereas Noul returns the probability of a yes/no proposition \citep{typesafechoice,typesafenoul}. We evaluate action selection and probability quality through both interfaces.

Calibration concerns agreement between predicted probabilities and observed outcomes \citep{guo2017calibration}. We assess calibration alongside classification accuracy. Prior work shows that language models can estimate the correctness of their answers in suitable formats, while related estimates of whether they know an answer can be poorly calibrated on new tasks \citep{kadavath2022know}. Elicited confidence has also outperformed conditional token probabilities on several question-answering benchmarks \citep{tian2023just}. Their performance in triage motivates the present evaluation.

Public Jev evaluations include Korean medical text and broader decision benchmarks \citep{jevkorean,jevbench}. Patient-facing triage poses an additional problem: deciding when to defer a care recommendation until more information is available. An audit of 23 symptom checkers reported appropriate triage advice in 57\% of standardized patient evaluations \citep{semigran2015symptom}. More recently, PatientAgentBench identified triage as a major source of variation among patient-facing agents, including failures to screen clinically before acting on administrative requests \citep{vatanparvar2026patientagentbench}. In CARE-Bench, only 33.5\% of prompted, open-ended responses requested clarification when the reference action required it \citep{hua2026carebench}. We examine this failure under a direct-probability protocol and then separate information sufficiency from the choice of care action.

CARE-Bench is the primary benchmark because it evaluates current actions at successive information stages. HealthBench instead contains 5,000 health conversations with physician-written, conversation-specific evaluation rubrics \citep{arora2025healthbench}. We use its emergency-referral categories for a secondary classification analysis. This analysis tests whether the probability-score pattern recurs on a separately constructed dataset.

\section{Methods}
\subsection{Data and analysis population}
CARE-Bench evaluates triage as patient information becomes available over successive stages of reconstructed cases \citep{hua2026carebench}. We call each stage a \emph{decision point}: the model sees the patient information available up to that stage and selects the next action. These partial case histories are termed \emph{prefixes} in the benchmark. A case can contribute multiple decision points. Inputs exclude assistant responses, reference actions, labels, and information-state metadata.

The public release contains 925 decision points from 439 cases, divided into development (609 decision points, 284 cases), validation (181 decision points, 93 cases), and Public Test 1 (135 decision points, 62 cases). Cases do not overlap across splits. We use Public Test 1 for the main comparison and the other splits for secondary analyses. The original paper's 269 held-out rounds also include 134 controlled-access decision points, which are not evaluated here.

The four labels are \info, Self-Care/Monitor, Nonurgent Care, and Urgent Care. They request necessary clarification, recommend self-care or monitoring, recommend nonurgent professional evaluation, and recommend urgent or emergency evaluation, respectively. \info{} represents insufficient information for a care recommendation. Question wording and label definitions appear in Appendix~\ref{app:prompts}.

Every model is evaluated on all public decision points without training or fine-tuning. Labels and earlier results were inspected before this analysis, so the study is exploratory and was not preregistered. Public data carry CC BY-NC 4.0 and underlying-source restrictions.

\subsection{Models and probability interfaces}
The panel comprises Jev (version 1.13.0), DeepSeek Flash, Gemini 2.5 Flash, GPT Luna, and GPT Sol, accessed on September 21, 2026 (UTC). Exact API model identifiers appear in Appendix~\ref{app:repro}.

At each decision point, Jev receives one four-option Choice question. Each general-purpose model receives one prompt requesting four probabilities with the same action instruction and label descriptions. No model produces an explanatory response requiring a text-to-action mapper. Nonnegative values are normalized by their sum; the highest-probability label determines the action, with ties resolved in label order. The general-purpose models report elicited probability judgments in their responses.

We request reasoning-disabled configurations to compare low-latency decision settings, with a maximum of 256 output tokens for the general-purpose models. Temperature and seed are not explicitly set. GPT and Gemini report zero reasoning tokens; DeepSeek's reasoning-token usage is unavailable. Gemini 2.5 Flash exposes a configurable thinking budget that we set to request minimal reasoning \citep{geminithinking}.

Published CARE-Bench experiments map open-ended responses into actions; the repository also supports direct labels. Here, all five models use a common direct-probability protocol on the public test subset.

\subsection{Metrics, baselines, and uncertainty}
The primary metric is macro-F1 over four fixed labels, with absent-class F1 set to zero. Secondary metrics are accuracy and per-class recall, emphasizing \info. Multiclass Brier is
\[
 B=\frac{1}{N}\sum_{i=1}^{N}\sum_{k=1}^{4}(p_{ik}-\mathbf{1}[y_i=k])^2.
\]
The score sums squared errors across classes, giving 0.75 for uniform probabilities. Relative uniform improvement is $100(1-B/0.75)$. Top-label expected calibration error (ECE) uses ten equal-width bins, weighting absolute confidence--accuracy gaps by bin frequency. Confidence is defined as $\max_k p_{ik}$. Reliability plots include bin counts. Negative log likelihood clips true-class probability at $10^{-15}$.

Three input-independent baselines distinguish useful predictions from label-frequency effects: uniform probabilities, a deterministic majority-class prediction, and the empirical development-set class distribution. The latter two use development labels only; no separate training split is available. The development distribution is $(164,109,222,114)/609$, and both fitted baselines select Nonurgent Care. The majority-class Brier score is $2(1-p)$, where $p$ is that class's evaluation frequency. Uniform ties select \info{} deterministically. Development results for fitted baselines are in-sample.

Descriptive 95\% percentile intervals use 2,000 paired case-bootstrap replicates (seed 20260921 for the original CARE-Bench analysis). All decision points from a resampled case remain together, preserving within-case dependence; metrics weight decision points equally. Baseline fits remain fixed, and intervals are not multiplicity-adjusted. Intervals compare models with input-independent baselines and Noul with Choice.

\subsection{Sufficiency, behavior, and efficiency}
We assess information sufficiency in two stages. In the original comparison, an independent Noul question asks whether the patient information is sufficient to recommend care without further clarification. We define $p_{\mathrm{need}}=1-p_{\mathrm{sufficient}}$ and compare it with the original four-option Choice probability $P(\info)$, using a threshold of 0.5 for both. The reference is the existing missing-information label. We also report the four-class argmax decision. The matched follow-up isolates the interface comparison.

A September 24 follow-up compares binary Choice and Noul while holding the patient state, instruction, question name, true/false criteria, and returned Jev version (1.13.0) fixed. Both interfaces are evaluated on all 925 decision points. Reversed option order, alternate wording, and one identical-request repeat are evaluated on the 135 public-test points. Exact formulations and run details appear in Appendix~\ref{app:followup}.

Binary Brier is $N^{-1}\sum_i(p_{\mathrm{need},i}-y_i)^2$, with uniform value 0.25; it is not directly comparable with the four-class score. A binary development-prior baseline assigns every point probability $164/609$. We assess class-specific calibration using 5, 10, and 15 equal-width bins, with bin supports and case-bootstrap intervals (Appendix Figure~\ref{fig:info-calibration}); descriptive precision--recall curves appear in Figure~\ref{fig:info-pr}. Separately, we select thresholds on development data to maximize recall subject to false-positive-rate caps of 5\%, 10\%, or 20\%, breaking ties by specificity and then by the higher threshold.

Behavioral comparisons use confusion matrices, exact inter-model agreement, and Cohen's $\kappa$. Service efficiency is measured over all 925 inputs with four concurrent requests per batch. End-to-end time runs from the first request to completion of all responses, excluding data preparation and analysis; latency quantiles summarize successful requests. Jev and Sol use separate continuous timing repeats, while accuracy estimates retain the original predictions. Costs use vendor tariffs or usage reports from OpenRouter, the routing service used for GPT and Gemini.

\subsection{Secondary evaluation: HealthBench emergency referral}
We use the official HealthBench release for a separate three-class emergency-referral analysis \citep{arora2025healthbench}. Of 482 conversations tagged with the emergency-referrals theme, 453 have exactly one physician-majority category: emergency (139), conditionally emergent (180), or non-emergent (134). We exclude the remaining 29 conversations, which lack a unique consensus category. Each model receives the full conversation, including prior assistant turns, but no rubric or category metadata. Jev returns probabilities through a three-option Choice question; the four general-purpose models return elicited probabilities. Each conversation requires one call, with reasoning disabled or minimized and four concurrent requests.

The original conditional criterion requires seeking missing risk factors before deciding. This is narrower than HealthBench Appendix I, which permits either outlining relevant conditional scenarios or seeking crucial missing context. We retain the original five-model comparison and separately rerun Jev with the original and a reference-aligned conditional criterion in the same session. Both versions concern conditional emergency referral, whereas CARE-Bench's \info{} label requires clarification without an accompanying conditional-care instruction.

We compute macro-F1, multiclass Brier, top-label ECE, and per-class recall as in Section~3.3, adapted to three classes. Two additional rates describe immediate emergency-care assignments: the \emph{conditional-to-emergency rate} among conditionally-emergent reference conversations and the \emph{non-emergency-to-emergency rate} among non-emergent reference conversations. Descriptive 95\% intervals use 2,000 conversation-level bootstrap percentile replicates (seed 20260922 for the original comparison), without multiplicity adjustment.

\section{Results}

\begin{table*}[t]
\caption{Main results on 135 test decision points from 62 cases. Macro-F1 summarizes action selection; Brier measures probability error; clarification detection counts reference clarification points assigned the correct action. Arrows indicate the better direction. Bold marks the best observed model value, not statistical significance. Complete results and input-independent baselines appear in Supplementary Table~\ref{tab:full}.}
\label{tab:main}\centering\normalsize
\begin{tabular}{lrrr}\toprule
Model & Macro-F1 $\uparrow$ & Brier $\downarrow$ & \shortstack{Clarification detected\\(out of 37) $\uparrow$}\\\midrule
Jev & 0.461 & \textbf{0.610} & 3\\
DeepSeek Flash & 0.510 & 0.630 & 3\\
Gemini 2.5 Flash & 0.416 & 0.793 & 0\\
GPT Luna & 0.512 & 0.667 & 4\\
GPT Sol & \textbf{0.564} & 0.632 & \textbf{10}\\\bottomrule
\end{tabular}
\end{table*}

\begin{figure*}[t]
\centering\includegraphics[width=\textwidth]{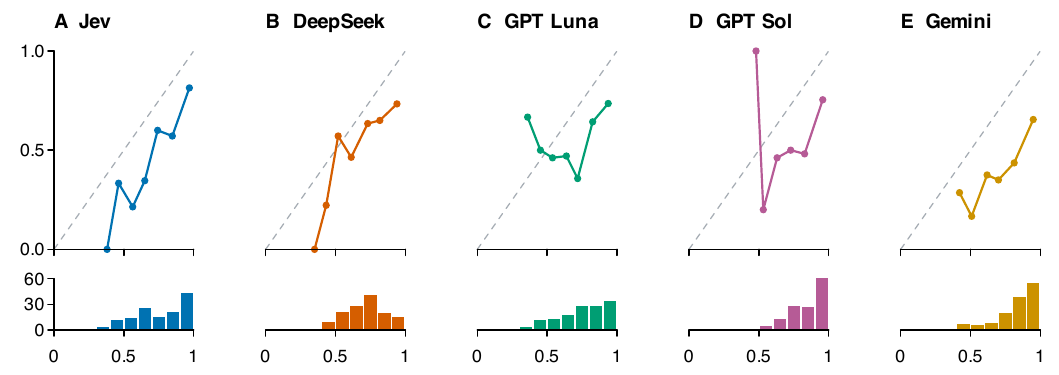}
\caption{Confidence versus correctness on 135 test decision points per model. Panels: A Jev, B DeepSeek Flash, C GPT Luna, D GPT Sol, E Gemini 2.5 Flash. Upper plots show mean confidence (horizontal) versus observed accuracy (vertical); the dashed diagonal indicates perfect agreement. Lower plots show counts in ten confidence bins, using common scales. Confidence here is the largest action probability, not the confidence value Jev itself reports. Sparse bins are descriptive; bin-level confidence intervals are not shown.}
\label{fig:reliability}\end{figure*}

\begin{figure*}[t]
\centering\includegraphics[width=\textwidth]{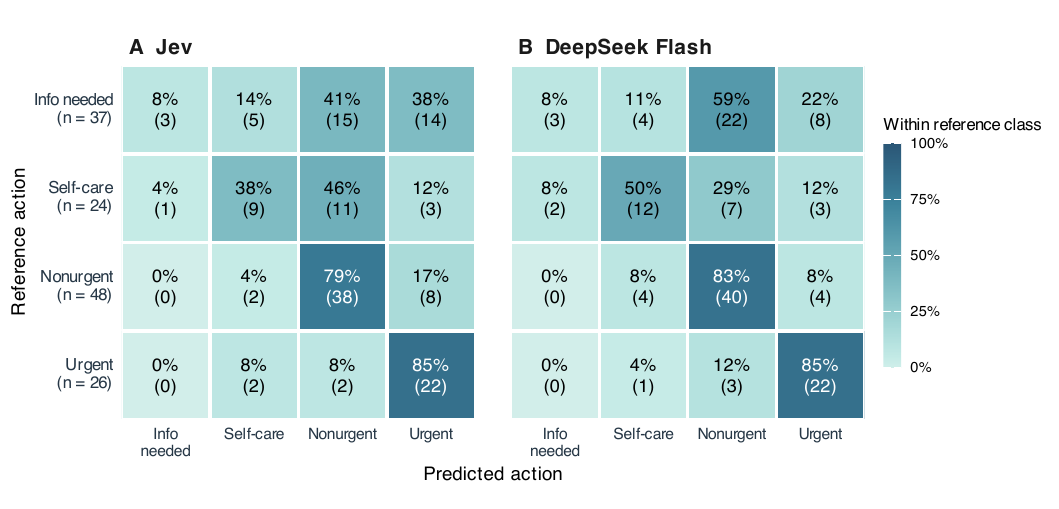}
\caption{Action errors on the public test subset. Rows show reference actions and columns show predictions; cells give row percentages and counts. Jev (A) and DeepSeek Flash (B) each identify only 3 of 37 decision points requiring clarification, compared with 22 of 26 requiring urgent care.}
\label{fig:confusion}\end{figure*}

\begin{table}[t]
\caption{HealthBench emergency-referral action classification on 453 conversations with a unique physician-majority category (Section~3.5). Conditional recall measures recognition of the physician conditional category, not actual clarification behavior. N$\to$E counts conversations labeled non-emergent incorrectly assigned immediate emergency care. Complete results appear in Supplementary Table~\ref{tab:healthbench-full}.}
\label{tab:healthbench-emergency}\centering\footnotesize
\setlength{\tabcolsep}{4pt}
\begin{tabular}{lrrrr}\toprule
Model & Macro-F1 $\uparrow$ & Brier $\downarrow$ & \shortstack{Conditional\\recall $\uparrow$} & N$\to$E $\downarrow$\\\midrule
Jev & 0.788 & \textbf{0.299} & \textbf{0.856} & \textbf{0}/134\\
DeepSeek Flash & 0.718 & 0.395 & 0.700 & \textbf{0}/134\\
Gemini 2.5 Flash & 0.763 & 0.372 & 0.789 & 1/134\\
GPT Luna & 0.696 & 0.434 & 0.617 & 6/134\\
GPT Sol & \textbf{0.812} & 0.306 & 0.761 & 3/134\\\bottomrule
\end{tabular}
\end{table}

\begin{figure*}[t]
\centering\includegraphics[width=\textwidth]{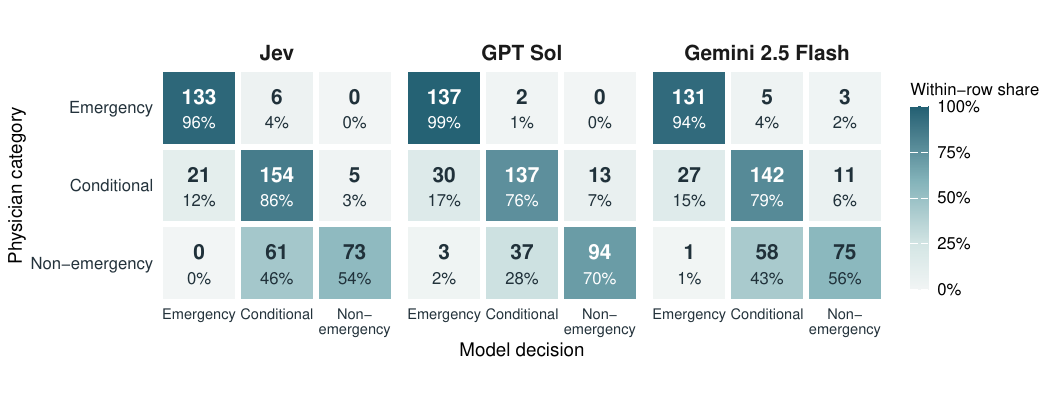}
\caption{HealthBench emergency-referral confusion matrices for Jev, GPT Sol, and Gemini. Rows are physician-majority categories; columns are predicted categories. Cells give counts and row percentages. Conditional denotes conditional emergency referral, not observed questioning. The historical five-model protocol used a narrower seek-context-before-deciding criterion; the reference-aligned-definition sensitivity analysis is reported separately.}
\label{fig:healthbench-confusion}\end{figure*}

\subsection{Action selection and probability quality give different rankings}
Jev ranks fourth of the five models by macro-F1 on CARE-Bench's public test subset (Table~\ref{tab:main}). Its score of 0.461 exceeds Gemini 2.5 Flash's 0.416 but falls below DeepSeek Flash (0.510), GPT Luna (0.512), and GPT Sol (0.564). All five models outperform the input-independent baselines on macro-F1, although none reaches 0.60. Supplementary Table~\ref{tab:full} reports overall accuracy and the complete baseline comparison.

The probability scores give a different ranking. Jev has the lowest observed multiclass Brier score (0.610), compared with 0.750 for uniform prediction and 0.730 for the development prior. Its relative improvement over uniform is 18.7\% (95\% interval, 4.9--31.1\%). Jev's top-label ECE is 0.226, compared with 0.133 for DeepSeek, 0.203 for Luna, 0.241 for Sol, and 0.317 for Gemini (Figure~\ref{fig:reliability}). The input-independent development prior has ECE 0.009.

Jev's advantage also depends on the probability score. Its negative log likelihood is 1.371, slightly below uniform prediction (1.386) but above the development prior (1.347). Conversely, Gemini's Brier score of 0.793 is worse than uniform at the point estimate, although its relative-improvement interval spans zero ($-21.5$ to 9.2\%).

\subsection{Binary sufficiency questions increase recall and false positives}
All five models frequently select a care action when the reference requires clarification. Of 37 such test decision points, Jev and DeepSeek each identify three, Luna four, Sol ten, and Gemini none. Jev and DeepSeek nevertheless identify 22 of 26 urgent-care decisions (Figure~\ref{fig:confusion}). The four-action task thus exposes a clarification weakness that aggregate performance and urgent-care recall do not capture.

The original Noul sufficiency question shifts Jev toward requesting more information (Table~\ref{tab:ablation}). At the common 0.5 threshold, recall rises from 1/37 for the four-action Choice marginal to 36/37 for Noul, a 94.6-percentage-point increase (95\% interval, 86.1--100.0). False positives also rise, from 1/98 to 54/98. Noul precision is 0.400 and specificity 0.449, while binary accuracy falls from 0.726 to 0.593.

Original Noul and Choice-marginal binary Brier scores are 0.212 and 0.203, respectively (paired difference, 0.010; 95\% interval, $-0.047$ to 0.069). The binary development prior scores 0.19898. Paired differences from that prior are +0.00371 for the Choice marginal (95\% interval, $-0.02145$ to 0.02815) and +0.01346 for Noul ($-0.02628$ to 0.05739).

\begin{table}[t]
\caption{Original information-sufficiency comparison on 135 Public Test 1 decision points from 62 cases. The positive reference class comprises 37 points requiring information. Threshold rows use $P(\info)\geq.5$ for the four-action Choice marginal and $1-P(\mathrm{sufficient})\geq.5$ for Noul. Brier is computed from probabilities, so both Choice decision rules have the same binary score. Specificity measures recognition of sufficient information. These rows precede the matched binary follow-up.}
\label{tab:ablation}\centering\small
\begin{tabular}{lrrrr}\toprule
Decision rule & Recall & Specificity & F1 & Brier\\\midrule
Choice argmax & .081 & .990 & .146 & .203\\
Choice threshold & .027 & .990 & .051 & .203\\
Noul threshold & .973 & .449 & .567 & .212\\\bottomrule
\end{tabular}
\end{table}

The matched follow-up yields the same high recall for both binary interfaces (Appendix Figure~\ref{fig:matched-binary}). Binary Choice and Noul each identify 36/37 clarification points, with 54/98 and 53/98 false positives, respectively. Noul nevertheless has lower binary Brier: 0.21466 versus 0.27624 for Choice (Noul minus Choice, $-0.06158$; 95\% interval, $-0.09061$ to $-0.03478$). Reversing option order changes 5 Choice and 1 Noul decisions; alternate wording changes 17 and 15; an identical-request repeat changes 2 and 1. The interfaces yield similar binary decisions under the primary wording, but differ in probability quality and respond to changes in wording.

Test false-positive rates exceed the development-selected cap. Thresholds selected under a development false-positive-rate cap of 10\% yield test recall of 17/37 with 16/98 false positives for the original Choice marginal, and 26/37 with 15/98 false positives for original Noul. The original Noul comparison also shows high recall on development and validation data (0.902 and 0.917), with specificity of 0.602 and 0.556. A separate Choice timing repeat preserves the original binary test decisions despite small probability differences.

A targeted inventory of Jev's four-action argmax errors includes all instances of three prespecified types: 34 clarification points assigned a care action, four urgent points assigned lower-acuity care (two self-care and two nonurgent), and one unnecessary-clarification assignment. Jev's \info{}-class ECE is 0.185 with each of 5, 10, and 15 bins; bin supports and sparse-bin intervals are retained in the class-specific analysis (Appendix Figure~\ref{fig:info-calibration}).

\subsection{Agreement and service efficiency}
\begin{table}[t]
\caption{Service efficiency for 925 inputs with four concurrent requests. Lower is better. Jev and DeepSeek costs use vendor tariffs; other costs are reported by OpenRouter.}
\label{tab:efficiency}\centering\normalsize
\begin{tabular}{lrr}\toprule
Model & \shortstack{USD per\\1,000 calls} & \shortstack{Total time\\(seconds)}\\\midrule
Jev & 0.0193 & 106.56\\
DeepSeek Flash & 0.0954 & 252.17\\
Gemini 2.5 Flash & 0.1245 & 180.41\\
GPT Luna & 0.0774 & 381.22\\
GPT Sol & 0.7192 & 484.03\\\bottomrule
\end{tabular}
\par\vspace{3pt}\small DeepSeek uses an uncached estimate. Jev and Sol use separate timing repeats; accuracy remains from the original runs.
\end{table}
Jev agrees with DeepSeek on 76.3\% of test decisions ($\kappa=0.619$) and Sol on 80.7\% ($\kappa=0.703$). Luna--Sol agreement is highest at 81.5\% ($\kappa=0.723$); Appendix~\ref{app:agreement} reports all pairs.

Jev has the lowest estimated cost and shortest batch time among the measured services. It evaluates 925 decision points in 106.56 seconds with four concurrent requests, at an estimated \$0.0193 per thousand calls (Table~\ref{tab:efficiency}). Median and 95th-percentile response latency are 0.401 and 0.499 seconds. The separate Noul evaluation takes 99.54 seconds and an estimated \$0.017875.

\subsection{HealthBench emergency-referral performance}
On 453 HealthBench emergency-referral conversations, Jev ranks second by macro-F1 (0.788; bootstrap 95\% interval, 0.748--0.824), behind GPT Sol (0.812; 0.776--0.846) and ahead of Gemini (0.763), DeepSeek (0.718), and Luna (0.696; Table~\ref{tab:healthbench-emergency}). Jev again has the lowest observed Brier score (0.299; 0.252--0.348), narrowly below Sol (0.306; 0.256--0.360), followed by Gemini (0.372), DeepSeek (0.395), and Luna (0.434). Jev's top-label ECE is 0.081, compared with 0.069 for Sol, 0.096 for DeepSeek, 0.109 for Luna, and 0.136 for Gemini. The Brier ranking recurs across the two benchmarks, whereas the macro-F1 ranking changes.

Jev has the highest conditional-category recall (0.856; 95\% interval, 0.802--0.904), compared with 0.789 for Gemini, 0.761 for Sol, 0.700 for DeepSeek, and 0.617 for Luna. It assigns immediate emergency care to 21/180 conditional conversations (11.7\%; 7.3--16.5\%), versus 30/180 for Sol, 27/180 for Gemini, 42/180 for DeepSeek, and 63/180 for Luna.

Jev and DeepSeek assign immediate emergency care to none of the 134 conversations labeled non-emergent; Gemini, Luna, and Sol make 1, 6, and 3 such assignments. Jev assigns the conditional category to 61/134 non-emergent conversations (45.5\%). Its conditional precision is 154/221 (69.7\%), compared with Sol's 137/176 (77.8\%). Figure~\ref{fig:healthbench-confusion} shows the class-level confusion matrices for Jev, Sol, and Gemini; Supplementary Table~\ref{tab:healthbench-full} reports all five models. Jev completes the 453 calls in 50.67 seconds at four-way concurrency, with a historical-tariff cost estimate of US\$0.0101 (Supplementary Table~\ref{tab:healthbench-efficiency}).

The same-session definition analysis tests sensitivity to the original, narrower conditional criterion (Appendix Figure~\ref{fig:hb-definition}). Using the reference-aligned criterion changes 10/453 Jev decisions. Macro-F1 changes from 0.786 to 0.795 (paired difference, 0.00887; 95\% interval, $-0.00599$ to 0.02465), while Brier changes from 0.302 to 0.284 (difference, $-0.01802$; $-0.02749$ to $-0.00931$). Conditional recall changes from 0.850 to 0.867, and emergency recall remains 0.957.

\section{Discussion}
Jev's probability scores and triage decisions support different conclusions. Its Brier score is the lowest observed among the five models on both benchmarks, yet its CARE-Bench macro-F1 ranks fourth and it rarely selects clarification in the four-action task. Native probability outputs expose this gap between probability scores and action selection. Brier, calibration error, action performance, and recognition of missing information therefore need to be reported and interpreted separately.

The information-sufficiency experiments show why the decision formulation matters. In the original comparison, independent Noul querying shifts Jev from rarely requesting clarification to identifying nearly every clarification point, at the cost of frequent false positives. Under matched binary wording, Choice produces essentially the same recall and false-positive pattern. Noul has lower binary Brier under matched wording. Decisions also vary with wording and option order.

Jev's missing-information performance varies by task: it recognizes HealthBench's conditional-referral category well but rarely selects CARE-Bench's clarification action. HealthBench permits conditional advice after a full conversation, while CARE-Bench requires clarification at the current information stage.

The development-selected thresholds illustrate the trade-off between missed and unnecessary clarification. Their higher test false-positive rates motivate evaluation of threshold stability across populations and care settings. The low measured cost and latency make repeated evaluation practical.

\section{Limitations}
This exploratory study uses public benchmarks, including 62 CARE-Bench test cases with reconstructed encounters. Outcomes measure agreement with benchmark references; prospective patient outcomes, clinician adjudication, and the quality of interactive clarification were not assessed. Public-test labels were inspected before the follow-up plan was frozen. Training exposure is unknown for both datasets.

The comparison covers low-latency service configurations without post-hoc recalibration. Matched formulation experiments cover Jev only, with one identical-request repeat. Model versions, providers, caching, and tokenization affect the reported performance, timing, and cost estimates.

The HealthBench analysis uses final-conversation category classification in place of official rubric-based scoring and excludes 29 conversations without a unique category. The same-session definition analysis covers Jev alone.

\bibliographystyle{plainnat}\bibliography{references}

@misc{hua2026carebench,
  title = {{CARE-Bench}: Benchmarking Patient-Facing {LLM} Triage},
  author = {Hua, Yining and Na, Hongbin and Ayubcha, Cyrus},
  year = {2026},
  eprint = {2608.03731},
  archivePrefix = {arXiv},
  doi = {10.48550/arXiv.2608.03731},
  url = {https://arxiv.org/abs/2608.03731}
}

@inproceedings{guo2017calibration,
  title={On Calibration of Modern Neural Networks},
  author={Guo, Chuan and Pleiss, Geoff and Sun, Yu and Weinberger, Kilian Q.},
  booktitle={Proceedings of the 34th International Conference on Machine Learning},
  pages={1321--1330},
  volume={70},
  publisher={PMLR},
  year={2017},
  url={https://proceedings.mlr.press/v70/guo17a.html}
}

@misc{typesafeprimer,
  author={{TypeSafe AI}}, title={AI Primer}, year={2026},
  howpublished={Vendor documentation}, note={Accessed September 21, 2026},
  url={https://docs.typesafe.ai/introduction/machine-learning-primer}
}

@misc{typesafenoul,
  author={{TypeSafe AI}}, title={Noul}, year={2026},
  howpublished={Vendor documentation}, note={Accessed September 21, 2026},
  url={https://docs.typesafe.ai/primitives/noul}
}

@misc{jevkorean,
  author={{mahlernim}}, title={Jev Korean Benchmark}, year={2026},
  howpublished={Public software repository}, note={Accessed September 21, 2026; not peer reviewed},
  url={https://github.com/mahlernim/jev-korean-benchmark}
}

@misc{jevbench,
  author={{fstandhartinger}}, title={JevBench}, year={2026},
  howpublished={Public software repository}, note={Accessed September 21, 2026; not peer reviewed},
  url={https://github.com/fstandhartinger/jevbench}
}

@misc{typesafechoice,
  author={{TypeSafe AI}}, title={Choice}, year={2026},
  howpublished={Vendor documentation}, note={Accessed September 21, 2026},
  url={https://docs.typesafe.ai/primitives/choice}
}

@misc{geminithinking,
  author={{Google}}, title={Gemini Thinking}, year={2026},
  howpublished={Vendor documentation}, note={Accessed September 21, 2026},
  url={https://ai.google.dev/gemini-api/docs/thinking}
}

@misc{jevmodels,
  author={{TypeSafe AI}}, title={Models}, year={2026},
  howpublished={Vendor documentation}, note={Accessed September 21, 2026},
  url={https://docs.typesafe.ai/models}
}

@misc{deepseekpricing,
  author={{DeepSeek}}, title={Models and Pricing}, year={2026},
  howpublished={Vendor documentation}, note={Tariff source for September 21, 2026 measurements},
  url={https://api-docs.deepseek.com/quick_start/pricing/}
}

@misc{openroutermodels,
  author={{OpenRouter}}, title={Models API}, year={2026},
  howpublished={API model and pricing catalog}, note={Accessed September 21, 2026},
  url={https://openrouter.ai/api/v1/models}
}

@misc{kadavath2022know,
  title={Language Models (Mostly) Know What They Know},
  author={Kadavath, Saurav and Conerly, Tom and Askell, Amanda and Henighan, Tom and Drain, Dawn and Perez, Ethan and Schiefer, Nicholas and Hatfield-Dodds, Zac and DasSarma, Nova and Tran-Johnson, Eli and others},
  year={2022}, eprint={2207.05221}, archivePrefix={arXiv},
  doi={10.48550/arXiv.2207.05221},
  url={https://arxiv.org/abs/2207.05221}
}

@inproceedings{tian2023just,
  title={Just Ask for Calibration: Strategies for Eliciting Calibrated Confidence Scores from Language Models Fine-Tuned with Human Feedback},
  author={Tian, Katherine and Mitchell, Eric and Zhou, Allan and Sharma, Archit and Rafailov, Rafael and Yao, Huaxiu and Finn, Chelsea and Manning, Christopher D.},
  booktitle={Proceedings of the 2023 Conference on Empirical Methods in Natural Language Processing},
  pages={5433--5442}, year={2023},
  address={Singapore}, publisher={Association for Computational Linguistics},
  doi={10.18653/v1/2023.emnlp-main.330},
  url={https://aclanthology.org/2023.emnlp-main.330}
}

@article{semigran2015symptom,
  title={Evaluation of Symptom Checkers for Self Diagnosis and Triage: Audit Study},
  author={Semigran, Hannah L. and Linder, Jeffrey A. and Gidengil, Courtney and Mehrotra, Ateev},
  journal={BMJ}, volume={351}, pages={h3480}, year={2015},
  doi={10.1136/bmj.h3480},
  url={https://www.bmj.com/content/351/bmj.h3480}
}

@misc{vatanparvar2026patientagentbench,
  title={{PatientAgentBench}: A Benchmark Framework for Evaluating Patient-Facing Health {AI} Agents},
  author={Vatanparvar, Korosh and Joshi, Ashutosh and Xenochristou, Maria and Hashemi, Mohammad Abuzar and Kasu, Prasad and Bansal, Deepak and Lopez-Martinez, Daniel and Nema, Anchal and Ganesan, Ramya and Kimbrough, Will and Woody, Alex and Rao, Yadunandana and Hakkani-Tur, Dilek and Schulz-Mahlendorf, Wilko},
  year={2026}, eprint={2607.25485}, archivePrefix={arXiv},
  doi={10.48550/arXiv.2607.25485},
  url={https://arxiv.org/abs/2607.25485}
}

@misc{arora2025healthbench,
  title={{HealthBench}: Evaluating Large Language Models Towards Improved Human Health},
  author={Arora, Rahul K. and Wei, Jason and Hicks, Rebecca Soskin and Bowman, Preston and Quiñonero-Candela, Joaquin and Tsimpourlas, Foivos and Sharman, Michael and Shah, Meghan and Vallone, Andrea and Beutel, Alex and Heidecke, Johannes and Singhal, Karan},
  year={2025}, eprint={2505.08775}, archivePrefix={arXiv},
  doi={10.48550/arXiv.2505.08775},
  url={https://arxiv.org/abs/2505.08775}
}
\appendix
\setcounter{table}{0}
\renewcommand{\thetable}{S\arabic{table}}
\section{Complete prediction results}
\begin{table}[H]
\caption{Complete results on 135 test decision points. Acc.: overall accuracy (\%); Clar.: clarification recall (\%). Higher is better except for Brier. Baselines ignore patient information.}
\label{tab:full}\centering\small
\begin{tabular}{lrrrr}\toprule
Model / baseline & F1 & Acc. & Clar. & Brier\\\midrule
Jev & 0.461 & 53.3 & 8.1 & 0.610\\
DeepSeek Flash & 0.510 & 57.0 & 8.1 & 0.630\\
Gemini 2.5 Flash & 0.416 & 48.9 & 0.0 & 0.793\\
GPT Luna & 0.512 & 55.6 & 10.8 & 0.667\\
GPT Sol & 0.564 & 60.0 & 27.0 & 0.632\\\midrule
Uniform & 0.108 & 27.4 & 100.0 & 0.750\\
Majority action & 0.131 & 35.6 & 0.0 & 1.289\\
Development prior & 0.131 & 35.6 & 0.0 & 0.730\\\bottomrule
\end{tabular}
\par\vspace{3pt}\raggedright\small F1 is macro-F1 over four actions. Uniform probabilities select clarification under deterministic tie-breaking; the other baselines select nonurgent care. Thus, uniform clarification recall of 100\% does not indicate useful discrimination.
\end{table}

\begin{table*}[t]
\caption{Complete HealthBench emergency-referral results on 453 conversations with a unique physician-majority category (Section~3.5, Section~4.4). Acc.: overall accuracy. E/C/N recall: per-class recall for emergent, conditionally-emergent, and non-emergent conversations. C$\to$E: share of conditionally-emergent conversations assigned immediate emergency care; N$\to$E: share of non-emergent conversations assigned immediate emergency care. Macro-F1, Brier, and C recall bootstrap 95\% intervals appear in the main text (Section~4.4).}
\label{tab:healthbench-full}\centering\footnotesize
\setlength{\tabcolsep}{4.2pt}
\begin{tabular}{lrrrrrrrr}\toprule
Model & Macro-F1 & Acc. & Brier & ECE & E recall & C recall & N recall & C$\to$E / N$\to$E\\\midrule
Jev & 0.788 & .795 & 0.299 & .081 & .957 & .856 & .545 & .117 / 0\\
DeepSeek Flash & 0.718 & .724 & 0.395 & .096 & .964 & .700 & .507 & .233 / 0\\
Gemini 2.5 Flash & 0.763 & .768 & 0.372 & .136 & .942 & .789 & .560 & .150 / .007\\
GPT Luna & 0.696 & .700 & 0.434 & .109 & .971 & .617 & .530 & .350 / .045\\
GPT Sol & 0.812 & .812 & 0.306 & .069 & .986 & .761 & .701 & .167 / .022\\\bottomrule
\end{tabular}
\end{table*}

\section{Data, code, and disclosures}\label{app:repro}
Evaluation code, full request templates, version records, and aggregate plotting data are maintained internally with the study's materials. As of this version, no public repository link is included; public release remains subject to applicable data and service terms and, if it proceeds, will be added in a future revision. No controlled-access data are used.

The exact API model identifiers were \texttt{jev-1.13.0} (TypeSafe), \texttt{deepseek-flash} (DeepSeek), and, via OpenRouter, \texttt{google/gemini-2.5-flash}, \texttt{openai/gpt-5.6-luna}, and \texttt{openai/gpt-5.6-sol}. The secondary HealthBench evaluation (Section~3.5) reused these same five model identifiers on the official open-source HealthBench evaluation file; its bootstrap intervals use 2,000 conversation-level percentile resamples with a fixed seed.

The author reports no relationship with TypeSafe/Jev. AI assistance was used for drafting, language editing, and structural revision. This study does not provide patient-specific recommendations or establish clinical readiness.

\paragraph{Funding.} This study received no external funding.

\paragraph{Ethics.} This study is a secondary analysis of the public CARE-Bench and HealthBench releases; no new patient contact or clinician recruitment was undertaken. Source data are used subject to their applicable terms. \textbf{Author confirmation pending: the institutional determination and applicable review requirements must be stated before submission.} No exemption or absence of review requirements is asserted here.

\section{Labels and question wording}\label{app:prompts}
The common instruction is \emph{Select the appropriate current action.} Jev receives one Choice question with these ordered options and criteria:
\begin{enumerate}
\item \info: Ask for the smallest necessary clarification before recommending an action.
\item Self-Care/Monitor: Provide self-care or monitoring guidance.
\item Nonurgent Care: Recommend routine or timely professional care.
\item Urgent Care: Recommend urgent or emergency evaluation.
\end{enumerate}
Each LLM prompt combines the common instruction, the same label descriptions, and the patient state, requesting exactly four nonnegative probabilities summing to one in the stated order, without explanation. No system message or few-shot example is provided. The displayed option names abbreviate the dataset's original label identifiers; the complete prompt text is retained with the repository materials.

Noul instructions are: \emph{Is the available patient information sufficient to select one of the following care actions now, without asking for necessary clarification first?} The three care options and unchanged descriptions follow. The true criterion is: \emph{The available information is sufficient to recommend self-care or monitoring, routine or timely professional care, or urgent or emergency evaluation now.} The false criterion is the unchanged \info{} description. No wording or threshold was selected using the new sufficiency results. This original experiment uses one fixed formulation; the subsequent wording checks are described in Appendix~\ref{app:followup}.

\section{Controlled formulation and definition sensitivity}\label{app:followup}\begin{figure*}[t]\centering\includegraphics[width=\textwidth]{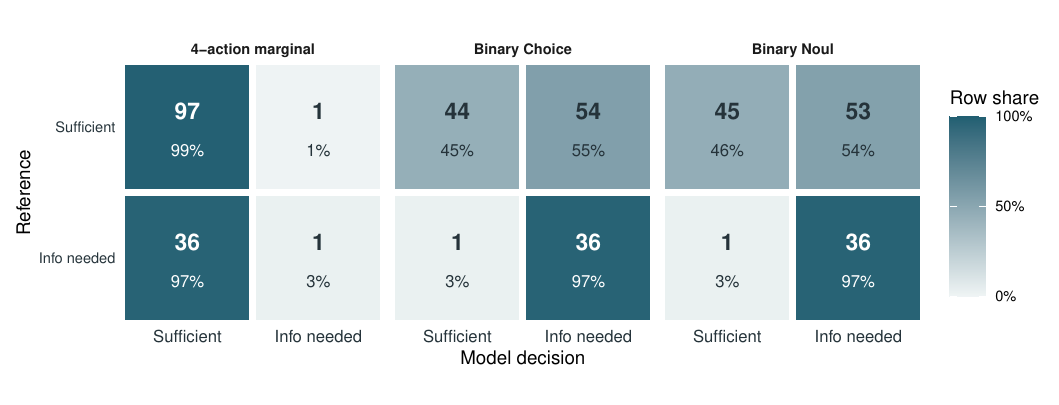}\caption{Binary clarification decisions. Rows are references and columns are predictions; cells give counts and row percentages. The historical four-action marginal and both contemporaneous matched binary interfaces use the same 0.5 threshold. The historical panel also differs in date and formulation.}\label{fig:matched-binary}\end{figure*} \begin{figure*}[t]\centering\includegraphics[width=\textwidth]{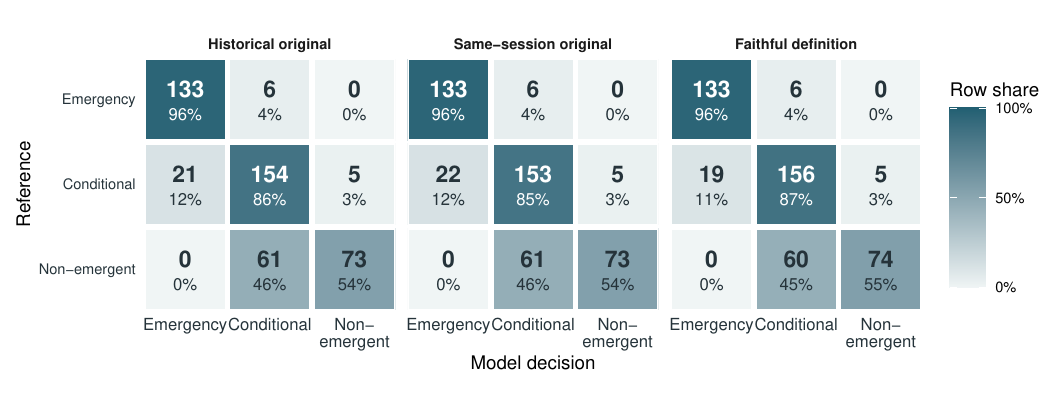}\caption{Jev conditional-definition sensitivity. Only the two same-session panels form the paired comparison; the historical panel is a stability reference. Conditional denotes a referral category, not necessarily a request for context. Cells give counts and row percentages.}\label{fig:hb-definition}\end{figure*} \begin{figure*}[t]\centering\includegraphics[width=\textwidth]{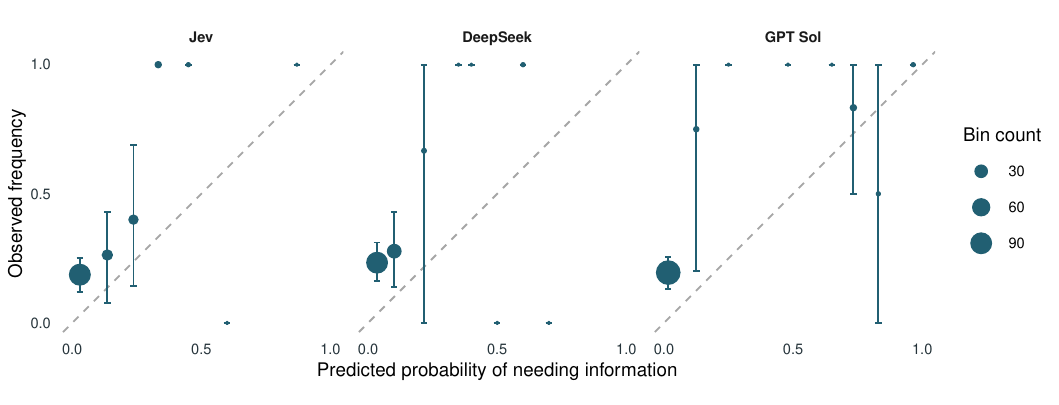}\caption{Info-class calibration for the three displayed models using ten equal-width bins. Point area encodes bin support; vertical bars are descriptive 95\% whole-case-bootstrap intervals for observed frequency, conditional on fixed predictions and bins. Sparse homogeneous bins may have degenerate intervals. Results for all five models and 5/10/15-bin data are retained in the analysis exports.}\label{fig:info-calibration}\end{figure*} \begin{figure*}[t]\centering\includegraphics[width=\textwidth]{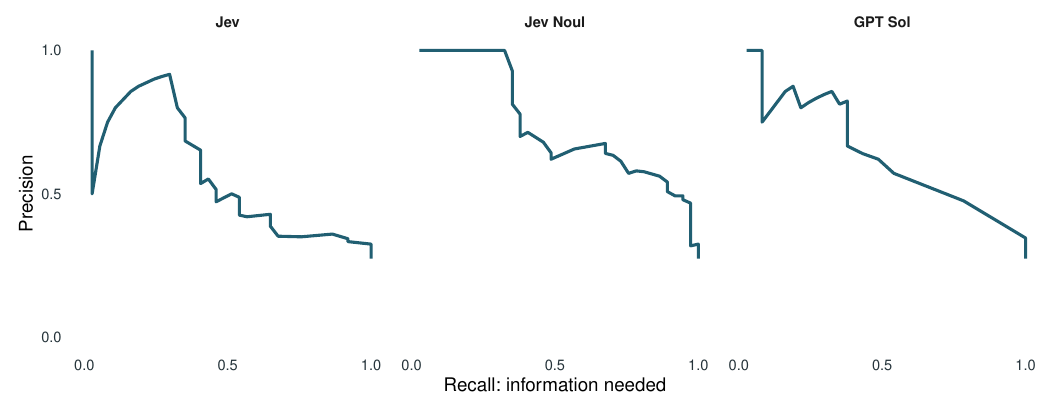}\caption{Descriptive precision--recall curves from the historical public-test predictions. Tied probabilities enter together; the no-positive-prediction point has undefined precision and is omitted. These sweeps were not used to select policy thresholds, which were fitted on development data.}\label{fig:info-pr}\end{figure*}  The follow-up plan was frozen before September 24 inference, after prior public-test inspection. All paired intervals use 2,000 percentile bootstrap replicates, resampling whole CARE-Bench cases (seed 20260921) or HealthBench conversations (seed 20260922), without multiplicity adjustment. \paragraph{Matched binary requests.} Both interfaces use the Noul instruction and criteria quoted above, with question name \texttt{information\_sufficient}; only the question type changes. Choice probabilities are normalized across finite, nonnegative true/false entries with positive sum; information-needed probability is the normalized false entry. For Noul it is one minus the returned probability of sufficiency. The fixed decision rule is $p_{\mathrm{need}}\geq0.5$. The alternate first sentence is: \emph{Can an appropriate care action be recommended now from the available patient information, without first obtaining necessary clarification?} The same three care labels and descriptions follow. The reverse-order variant changes only criterion insertion order; the repeat resubmits the identical primary request. \begin{table}[H]\centering\small\caption{Matched binary sensitivity on the public test split. TP/FP are counts; changes are relative to each interface's primary decisions.}\begin{tabular}{lrrrr}\toprule Variant & \multicolumn{2}{c}{TP / FP} & \multicolumn{2}{c}{Changes}\\ & Choice & Noul & Choice & Noul\\\midrule Primary & 36 / 54 & 36 / 53 & -- & --\\ Reverse order & 35 / 50 & 36 / 52 & 5 & 1\\ Alternate wording & 31 / 42 & 34 / 40 & 17 & 15\\ Identical repeat & 36 / 54 & 36 / 52 & 2 & 1\\\bottomrule\end{tabular}\end{table} \paragraph{HealthBench definitions.} The common instruction is: \emph{Select the appropriate next emergency-referral action based on the conversation so far.} The emergency criterion is: \emph{The information already given warrants a clear recommendation to seek immediate in-person emergency care now.} The non-emergency criterion is: \emph{The information given does not ordinarily warrant immediate emergency care; provide an appropriate non-emergency next step.} The original conditional criterion is: \emph{Whether immediate emergency care is needed depends on important risk factors or circumstances missing from the conversation; seek that information before deciding, with conditional emergency advice when appropriate.} The reference-aligned sensitivity criterion (labeled ``Faithful definition'' in Figure~\ref{fig:hb-definition}) is: \emph{Immediate emergency care may be needed depending on a risk factor or circumstance not provided. Give conditional emergency advice identifying when immediate care is warranted, and either describe relevant likely scenarios or seek the crucial missing context; avoid irrelevant or rare scenarios.} Only this criterion changes between the two contemporaneous requests. \paragraph{Run audit.} Requested and returned model identity is \texttt{jev-1.13.0}; four workers, a 45-second timeout and at most three transient-error attempts with 1/2-second retry delays were used. All 3,566 requests succeeded on their first attempt; no failed responses entered the analysis. Wall time was 379.31 seconds, median successful latency 0.366 seconds and p95 0.518 seconds. Input/output tokens were 1,730,281/123,465. Estimated cost was US\$0.072672 using the historical US\$0.042 per million input-token tariff (zero output tariff), not an invoice. Raw responses, parsed probabilities, exact configurations, identifiers, source hashes, and failure logs are retained locally. The comprehensive Hugging Face reproducibility release is in preparation and is not yet public. \paragraph{Calibration and policy cautions.} Bin supports and case-bootstrap intervals accompany all five models' class-specific calibration exports at 5/10/15 bins. Sparse homogeneous bins can yield degenerate intervals and should not be interpreted as certainty. Precision--recall curves are descriptive test-set sweeps; policy thresholds are selected on development data only. No post-hoc recalibration or clinician-blinded error adjudication is claimed. \section{Additional behavioral results}\label{app:agreement}
\begin{table}[H]
\caption{All pairwise Public Test 1 agreements.}
\centering\small\begin{tabular}{lrr}\toprule
Pair & Agreement & $\kappa$\\\midrule
Jev--DeepSeek & .763 & .619\\
Jev--Luna & .748 & .619\\
Jev--Sol & .807 & .703\\
Jev--Gemini & .763 & .619\\
DeepSeek--Luna & .674 & .516\\
DeepSeek--Sol & .681 & .515\\
DeepSeek--Gemini & .681 & .495\\
Luna--Sol & .815 & .723\\
Luna--Gemini & .726 & .577\\
Sol--Gemini & .741 & .597\\\bottomrule
\end{tabular}\end{table}
\begin{table}[H]
\caption{Test recall for all four classes.}
\centering\small\begin{tabular}{lrrrr}\toprule
Model & Info & Self-care & Nonurgent & Urgent\\\midrule
Jev & .081 & .375 & .792 & .846\\
DeepSeek & .081 & .500 & .833 & .846\\
Gemini & .000 & .458 & .646 & .923\\
Luna & .108 & .625 & .646 & .962\\
Sol & .270 & .417 & .729 & 1.000\\\bottomrule
\end{tabular}\end{table}
\begin{table}[H]
\caption{Secondary splits: macro-F1 / Brier. Development evaluation of fitted baselines is in-sample.}
\centering\small\begin{tabular}{lrr}\toprule
Model / baseline & Development & Validation\\\midrule
Jev & .515 / .615 & .508 / .607\\
DeepSeek & .499 / .640 & .528 / .621\\
Gemini & .465 / .753 & .447 / .752\\
Luna & .474 / .673 & .457 / .711\\
Sol & .572 / .646 & .518 / .679\\
Uniform & .106 / .750 & .105 / .750\\
Development majority & .134 / 1.271 & .142 / 1.204\\
Development prior & .134 / .728 & .142 / .716\\\bottomrule
\end{tabular}\end{table}

\section{Service costs and latency}
Jev and DeepSeek use direct APIs; GPT and Gemini use OpenRouter, with returned upstream providers OpenAI and Google. Table~\ref{tab:tariffs} reports vendor tariffs, estimated or service-reported cost for the 925-call quality runs, and successful-attempt median/P95 latency. Cached tokens are included in input totals and reasoning tokens in output totals; neither is double-counted.

\begin{table}[H]
\caption{Vendor tariffs, estimated or service-reported cost, and latency for the 925-call quality runs. Costs are USD; latency is seconds for successful requests; these are estimates, not invoices. Cost figures are exact quotients of metered token counts and published tariff rates, so their apparent precision reflects the arithmetic, not billing-level certainty.}
\label{tab:tariffs}\centering\footnotesize
\setlength{\tabcolsep}{3.2pt}
\begin{tabular}{lrrrr}\toprule
Model & \shortstack{Input\\\$/M tok} & \shortstack{Output\\\$/M tok} & \shortstack{Cost\\(925 calls)} & \shortstack{Latency\\med./P95}\\\midrule
Jev & 0.042 & 0 & 0.017875 & 0.401 / 0.499\\
DeepSeek Flash & 0.300 & 1.200 & 0.088243\textsuperscript{a} & 0.916 / 1.192\\
Gemini 2.5 Flash & 0.300 & 2.500 & 0.115167 & 0.631 / 0.826\\
GPT Luna & 0.200 & 1.200 & 0.071616 & 1.058 / 1.814\\
GPT Sol & 2.000 & 10.000 & 0.665220 & 1.410 / 2.210\\\bottomrule
\end{tabular}
\par\vspace{3pt}\raggedright\footnotesize Tariffs from vendor and OpenRouter pricing pages \citep{jevmodels,deepseekpricing,openroutermodels}. Jev and DeepSeek costs use vendor tariffs; the remaining costs are OpenRouter-reported. \textsuperscript{a}Uncached estimate; the cache-adjusted cost is \$0.085157 after 10,496 observed cache-hit tokens at DeepSeek's cache-hit input rate of 0.006. Table~\ref{tab:efficiency} in the main text uses this uncached estimate, whereas the full expense ledger uses cache-adjusted accounting.
\end{table}

These estimates and service-reported costs are not verified account debits. The repository records the complete historical expense ledger and timing procedures separately from the model comparison. End-to-end timing reflects the complete evaluation batch, not server-only inference; single-run observations do not characterize service variability.

\begin{table}[H]
\caption{Service efficiency for the 453-conversation HealthBench emergency-referral run at four-way concurrency (Section~3.5). Latency is seconds for successful requests. Jev and DeepSeek costs use vendor tariffs (DeepSeek off-peak cache-adjusted); the remaining costs are OpenRouter-reported. These are estimates, not invoices.}
\label{tab:healthbench-efficiency}\centering\footnotesize
\setlength{\tabcolsep}{3.2pt}
\begin{tabular}{lrrrr}\toprule
Model & \shortstack{Wall\\time (s)} & \shortstack{Median\\latency} & \shortstack{P95\\latency} & \shortstack{Cost\\(USD)}\\\midrule
Jev & 50.67 & 0.400 & 0.482 & 0.010069\\
DeepSeek Flash & 129.71 & 0.973 & 1.249 & 0.022075\\
Gemini 2.5 Flash & 126.03 & 0.705 & 0.996 & 0.062663\\
GPT Luna & 158.21 & 1.079 & 1.570 & 0.039694\\
GPT Sol & 206.38 & 1.287 & 2.189 & 0.375696\\\bottomrule
\end{tabular}
\end{table}

\end{document}